\documentclass[11pt, letter, twocolumn, goog]{google}

\usepackage[numbers]{natbib}
\uselogo{} 

\title{Agentic AI and the next intelligence explosion}

\renewcommand{\today}{}

\author[1,2,3]{James Evans}
\author[1,4,5]{Benjamin Bratton}
\author[1,3]{Blaise Ag\"uera y Arcas}

\affil[1]{Paradigms of Intelligence Team, Google}
\affil[2]{University of Chicago}
\affil[3]{Santa Fe Institute}
\affil[4]{Antikythera, Berggruen Institute}
\affil[5]{University of California, San Diego}

\begin{abstract}

\end{abstract}

\begin{document}

\maketitle
\noindent For decades, the artificial intelligence (AI) ``singularity''~\citep{vinge1993} has been heralded as a single, titanic mind bootstrapping itself to godlike intelligence~\citep{kurzweil2005}, consolidating all cognition into a cold silicon point. But this vision is almost certainly wrong in its most fundamental assumption. If AI development follows the path of previous major evolutionary transitions~\citep{szathmary1995} or ``intelligence explosions''~\citep{aguera2025}, our current step-change in computational intelligence will be plural, social, and deeply entangled with its forebears (us!).

By its nature, intelligence is high-dimensional and relational, not a single quantity that must be unambiguously less or greater than human scale. In fact, it is unclear what we even mean by ``human scale,'' given that our intelligence is already a collective property, not an individual one. Recent advances in agentic AI show us once again that intelligence has always fundamentally involved the interaction of distinctive, distributed perspectives~\citep{woolley2010}, and it is from social organization~\citep{mercier2011} that transformative intelligence has and will continue to emerge.

We can observe this in at least two ways: In the orchestration of societies of AI agents~\citep{aguera2026silicon} by and with human users in new ``centaur'' configurations, and in the microsocieties that flourish inside and between reasoning models themselves. Let's start with the latter.

What happens inside an ostensibly singular reasoning model? A community conversation, as it turns out. In a recent study, we demonstrated that frontier reasoning models like DeepSeek-R1 and QwQ-32B do not improve simply by ``thinking longer.'' Instead, they simulate complex, multi-agent-like interactions within their own chain of thought---what we term a ``society of thought''~\citep{kim2026}. These models spontaneously generate internal debates among distinct cognitive perspectives that argue, question, verify, and reconcile. This conversational structure causally accounts for the models' accuracy advantage on hard reasoning tasks, which we demonstrated by explicitly priming and amplifying multi-party conversation.

The finding is striking because it reveals an emergent behavior. None of these models were trained to produce societies of thought. When reinforcement learning is used to reward base models solely for reasoning accuracy, they spontaneously increase conversational, multi-perspective behaviors~\citep{aguera2026emergent}. Models are rediscovering, through optimization pressure alone, what centuries of epistemology and decades of cognitive science~\citep{mercier2017} have suggested: that robust reasoning is a social process~\citep{moshman1986}, even when it occurs within a single mind~\citep{mead2015}. The exact nature of that emergent behavior is, of course, to be further discovered (and invented) as cooperative human-agent social dynamics become more grounded, complex, and durable. What proves fundamental about socially-mediated reasoning in general, and what is specific to fine-tuned and reinforced contexts, is likely to inspire considerable research in the coming years.

This opens a vast---yet familiar---design space. The social and organizational sciences have spent a century studying how team size~\citep{wuchty2007}, composition, hierarchy~\citep{xu2022}, role differentiation, conflict norms, institutions, and network structures shape collective performance. Almost none of this research has been brought to bear on AI reasoning~\citep{xu2025}. Today's reasoning models produce a single conversation---an AI town hall transcript. But effective groups exhibit hierarchy, specialization, division of labor, and structured disagreement. To explore this, we will need systems that support multiple parallel, converging, and diverging streams of deliberation---architectures in which brainstorming, devil's advocacy, and constructive conflict are not accidental emergent properties but designed features. The toolkits of team science, small-group sociology, and social psychology become blueprints for next-generation AI development.

In addition to its practical applications, these insights may clarify the entire history of intelligence. Each prior ``intelligence explosion'' was not an upgrade to individual cognitive hardware, but the emergence of a new, socially aggregated unit of cognition~\citep{maynardsmith1997}. Primate intelligence scaled with social group size~\citep{dunbar1998}, not habitat difficulty. Human language created what Michael Tomasello calls the ``cultural ratchet''~\citep{tomasello1999}: knowledge accumulating across generations without any individual requirement to reconstruct the whole. Writing, law, and bureaucracy externalized social intelligence into infrastructure~\citep{goody1986}, institutions that coordinate across longer time horizons than any participant within them. A Sumerian scribe running a grain accounting system did not comprehend its macroeconomic function; the system was functionally more intelligent than he was.

AI extends this sequence. Large language models are trained on the accumulated output of human social cognition~\citep{omadagain2021}---the cultural ratchet made computationally active, every parameter a compressed residue of communicative exchange. What migrates into silicon is not abstract reasoning but social intelligence in externalized form~\citep{farrell2025}, encountering itself on a new substrate.

If intelligence is inherently social, then the path to more powerful AI runs not through building a single colossal oracle but through composing richer social systems---and these systems will be hybrid. We have entered the era of human-AI centaurs: composite actors that are neither purely human nor purely machine. Centaur actors can take many forms and inhabit many different roles. Each one of us may move in and out of diverse ensembles many times a day: one human directing many AI agents; one AI serving many humans; many humans and many AIs collaborating in shifting configurations~\citep{lovegrove2006}.

A corporation or state comprising myriad of humans already holds singular legal standing and acts with collective agency that no individual member can fully control. The recent explosion of agentic AI suggests the possibility of something similar at the scale of billions of interacting minds, human and non-human alike. Platforms like OpenClaw, an open source platform for building multi-purpose AI agents that persist within a computer, and Moltbook, a popular social network for AI agents to interact, offer embryonic glimpses~\citep{aguera2026silicon} of this future. But the deeper structural shift goes beyond any single platform. Agents can now renew and fork themselves, splitting into two versions, and interact with one another; an agent facing a complex task can initiate new copies, differentiate and assign them subtasks, then recombine the results. Imagine an agent confronting a dauntingly complex problem spawns an internal society of thought. One emergent perspective, encountering a sub-problem beyond its reach, spawns its own subordinate society, a recursive descent into collective deliberation that expands when complexity demands and collapses when the problem resolves. Conflict is not a bug but a resource, flexibly instantiated and dissolved at every level of the folding and unfolding hypergraph of conversations.

This implies a very different approach to scaling. It is not only about scaling the raw computational capacity of an agent, but about building systems that can operate at the scale and within the context of a real society. This means putting as much effort into building agent institutions as building agents themselves. The dominant paradigm for AI alignment---Reinforcement Learning from Human Feedback~\citep{christiano2017}---resembles a parent-child model of correction, fundamentally dyadic and unable to scale to billions of agents. The social intelligence perspective suggests an alternative: institutional alignment~\citep{ostrom1990}. Just as human societies rely not on individual virtue but on persistent institutional templates~\citep{north1990}---courtrooms, markets, bureaucracies---defined by roles and norms, scalable AI ecosystems will require digital equivalents~\citep{bai2022}. The identity of any agent matters less than its ability to fulfill a role protocol, just as a courtroom functions because ``judge,'' ``attorney,'' and ``jury'' are well-defined slots, independent of who occupies them.

Nowhere is this more urgent than in governance itself. When AI systems are deployed in high-stakes decisions---hiring, sentencing, benefits allocation, regulatory enforcement---the question of who audits the auditors becomes unavoidable. The answer may be constitutional in structure. Governments will need AI systems with distinct, explicitly invested values---transparency, equity, due process---whose function is to check and balance AI systems deployed by the private sector and other branches of government, and vice versa. For example, a labor department AI may audit a corporation's hiring algorithm for disparate impact; a judicial branch AI may evaluate whether an executive branch AI's risk assessments meet constitutional standards. The alternative is, for example, for the U.S. Securities and Exchange Commission to ineffectively hire business school graduates armed with Excel spreadsheets to combat high-dimensional collusion of AI-augmented trading platforms.

``Governance,'' however, does not only mean what governments do. Governance systems, in the cybernetic sense of the term, need to be built into human-agent and agent-to-agent systems as they grow and complexify. This will likely entail means to ensure and verify outcomes and decisions of multiple-stakeholder deliberation, procedural delegation of tasks and sub-tasks and reliable scaffolds for automating delicate inter-agent collaborations. Such protocols may have as much real-world effect for ``agent governance'' as any laws will.

Crucially, humans remain in the loop. Agent institutions are populated by both humans and AI agents in different roles and configurations. It's not ``either/or'', but ``both/and''. The U.S. Founders would have recognized the logic~\citep{hamilton1788}: no single concentration of intelligence, human or artificial, should regulate itself. Power must check power, and in a world of artificial agents, this means building conflict and oversight into the institutional architecture.

The vision we describe is neither utopian nor dystopian; it is evolutionary. Any emergent intelligence explosion will be seeded by eight billion humans interacting with hundreds of billions, eventually trillions, of AI agents. The scaffold is not a single mind ascending but a combinatorial society complexifying: intelligence growing like a city~\citep{spencer1857}, not a single meta-mind.

A ``monolithic singularity'' framework leads to policies aimed at preventing a technology that may never exist. Instead, we should be looking for the next intelligence explosion in the same place from which the previous ones emerged: in cooperative, competitive and creative interaction between multitudes of socially intelligent minds. The difference this time is that most of those minds will be non-biological. This plurality model~\citep{weyl2024} focuses attention where it belongs: on the design of mixed human-AI social systems, the norms that govern them, and the institutions and protocols through which they conflict and coordinate.

In a very real sense, the intelligence explosion is already here~\citep{weyl2024}---in the society of thought debating inside every reasoning model, in the centaur workflows reshaping every knowledge profession, in the recursive agent ecologies beginning to fork and collaborate at scale, and in the constitutional questions we must now begin to ask. The question is not whether intelligence will become radically more powerful, but whether we will build the social infrastructure worthy of what it is becoming. No mind is an island.

\bibliography{main}

\end{document}